# Retrieval-Augmented Generation in Medicine: A Scoping Review of Technical Implementations, Clinical Applications, and Ethical Considerations

Rui Yang[1,2]†, Matthew Yu Heng Wong[3]†, Huitao Li[1,2]†, Xin Li[1,2], Wentao Zhu[1,2], Jingchi Liao[1,2], Kunyu Yu[1,2], Jonathan Chong Kai Liew[1,2], Weihao Xuan[4], Yingjian Chen[5], Yuhe Ke[2,6], Jasmine Chiat Ling Ong[2,7], Douglas Teodoro[8], Chuan Hong[9,10], Daniel Shi Wei Ting[11,12,13], Nan Liu[1,2,9,14,15]*

*†: co-first authors    \*: corresponding author*
*[1] Center for Quantitative Medicine, Duke–NUS Medical School, Singapore 169857, Singapore*
*[2] Duke-NUS AI + Medical Sciences Initiative, Duke-NUS Medical School, Singapore 169857, Singapore*
*[3] School of Clinical Medicine, University of Cambridge, Cambridge CB2 0SP, UK*
*[4] Graduate School of Frontier Sciences, The University of Tokyo, Tokyo 277-8561, Japan*
*[5] Graduate School of Engineering, The University of Tokyo, Tokyo 113-8654, Japan*
*[6] Division of Anesthesiology and Perioperative Medicine, Singapore General Hospital, Singapore 169608, Singapore*
*[7] Division of Pharmacy, Singapore General Hospital, Singapore 169608, Singapore*
*[8] Department of Radiology and Medical Informatics, University of Geneva, Geneva 1202, Switzerland*
*[9] Department of Biostatistics and Bioinformatics, Duke School of Medicine, Durham, NC 27710, USA*
*[10] Duke Clinical Research Institute, Durham, NC 27705, USA*
*[11] Singapore Eye Research Institute, Singapore National Eye Center, Singapore 168751, Singapore*
*[12] Byers Eye Institute, Stanford University, Stanford, CA 94303, USA*
*[13] Artificial Intelligence Office, Singapore Health Services, Singapore 168582, Singapore*
*[14] Pre-hospital & Emergency Research Centre, Health Services and Systems Research, Duke–NUS Medical School, Singapore 169857, Singapore*
*[15] NUS Artificial Intelligence Institute, National University of Singapore, Singapore 119391, Singapore*

\*Corresponding Author: Nan Liu, Centre for Quantitative Medicine, Duke-NUS Medical School, 8 College Road, Singapore 169857, Singapore
Email: liu.nan@duke-nus.edu.sg

## Summary

The rapid growth of medical knowledge and increasing complexity of clinical practice pose challenges. In this context, large language models (LLMs) have demonstrated value; however, inherent limitations remain. Retrieval-augmented generation (RAG) technologies show potential to enhance their clinical applicability. This study reviewed RAG applications in medicine. We found that research primarily relied on publicly available data, with limited application in private data. For retrieval, approaches commonly relied on English-centric embedding models, while LLMs were mostly generic, with limited use of medical-specific LLMs. For evaluation, automated metrics evaluated generation quality and task performance, whereas human evaluation focused on accuracy, completeness, relevance, and fluency, with insufficient attention to bias and safety. RAG applications were concentrated on question answering, report generation, text summarization, and information extraction. Overall, medical RAG remains at an early stage, requiring advances in clinical validation, cross-linguistic adaptation, and support for low-resource settings to enable trustworthy and responsible global use.

## Keywords

large language models, retrieval-augmented generation, global health equity

# Introduction

Contemporary clinical practice is facing the rapid expansion of medical knowledge and the increasing complexity of diagnostic and therapeutic decision-making.[1] At the same time, the growing demand for personalized health care posts unprecedented challenges in the retrieval, integration, and application of medical information, further intensifying clinicians' workload.[2] In this context, large language models (LLMs) are gradually being introduced into medicine and have demonstrated potential value. Existing proprietary LLMs[3–13] and open-weight LLMs[14–24] have shown outstanding performance in generative tasks, while reasoning-oriented LLMs have further extended the boundaries of complex tasks.[25–28] Meanwhile, specific LLMs such as Med-Gemini[29,30] and MedGemma[31] have achieved high adaptability to medical scenarios by incorporating specific knowledge.[32] Collectively, these advances highlight the potential applications of LLMs in clinical consultation,[33] disease diagnosis,[34–37] treatment management,[33] medical education,[38] and scientific research,[39] while also suggesting their role in alleviating the burden on clinicians and improving overall health care quality.[32,33]

Despite this, LLMs face significant challenges in medical applications. First, LLMs rely on static training data, limiting their ability to keep pace with rapidly evolving medical knowledge.[40] Second, they are prone to generate content without factual grounding.[36] Third, LLMs typically function as "black boxes" with outputs lacking explainability.[36] Fourth, they cannot access private patient-specific data or hospital-specific guidelines, restricting their utility in personalized diagnostic and treatment support.[40] Lastly, LLMs may perpetuate inherent biases from training data, potentially exacerbating disparities for specific populations and further widening health inequities.[41] These issues are not only technical bottlenecks but also involve safety and ethical governance as well as global health equity.[42]

Against this backdrop, retrieval-augmented generation (RAG) has emerged as a promising solution.[40,43] RAG technologies enable LLMs to incorporate information from external sources during the generation process, providing outputs that are up-to-date, relevant, and fact-grounded. The initial "Naive RAG" follows an "index-retrieve-generate" pipeline, as shown in Figure 1. The RAG system first retrieves information from external knowledge sources, such as research literature or clinical guidelines, and then augments LLMs with the relevant information, helping them generate the answer.[40] Later, "Advanced RAG" introduces pre-retrieval optimization and post-retrieval processing strategies to improve the quality of retrieved content.[36] Meanwhile, "Modular RAG" provides a more flexible architecture, allowing for the combination of functional modules adapted to specific medical scenarios. These technological advancements offer different pathways to mitigate the limitations of LLMs.[36,44]

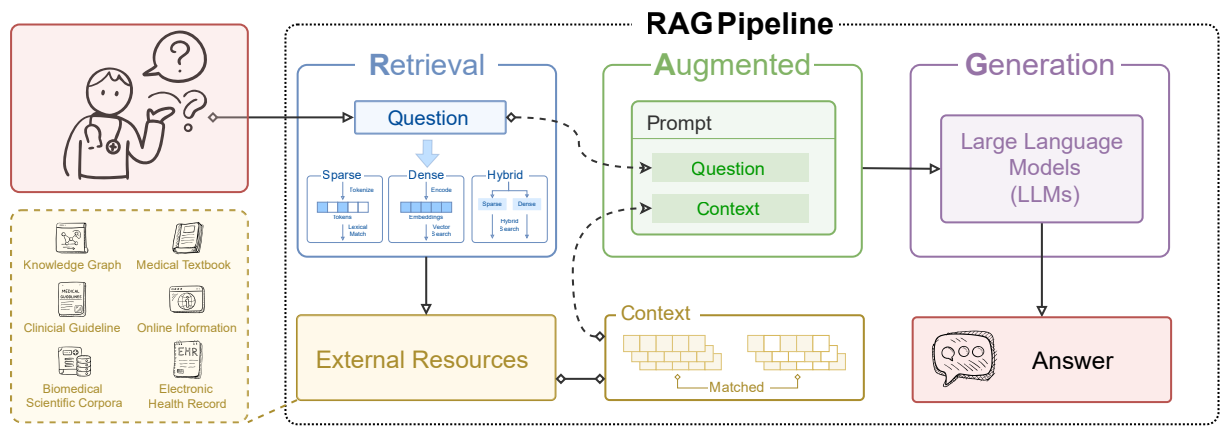

**Figure 1. Naive RAG framework.** The framework consists of three stages: first, the retrieval module obtains information relevant to the query; next, the query and the retrieved information are provided to the LLM; finally, the LLM generates the answer.

Several studies have examined the development of RAG in medicine, providing valuable perspectives on this rapidly evolving field. Yang et al. analyze the possible contributions that RAG could bring to health care in equity, reliability, and personalization.[40] Liu et al. conducted a meta-analysis of 20 studies, demonstrating that RAG systems improved the performance of LLMs, and proposed guidelines for unified implementation and development of enhanced LLM applications with RAG in clinical settings.[45] Amugongo et al. carried out a systematic review with a focus on methodology, analyzing 70 studies to compare different RAG paradigms and their technical implementations in medical scenarios.[46] He et al. surveyed RAG datasets, technologies, and applications in medicine, with particular emphasis on technical components and system architectures.[43] While existing studies offer important insights, they have not systematically mapped the implementation pathways and application patterns of RAG in medicine. More critically, the evaluation of RAG technologies—especially concerning bias, safety, and deployment in low-resource settings—remains insufficiently explored. However, these factors are crucial for advancing the equitable global application of RAG technologies.

This scoping review aims to systematically outline the research landscape of RAG in medicine, mapping its implementation pathways and application patterns, and evaluating its potential value in addressing the rapid evolution of medical knowledge and other critical dimensions of clinical practice. Additionally, we emphasize that while RAG alleviates certain limitations of LLMs and facilitates their applications in medicine, it simultaneously introduces challenges. These challenges go beyond the reliability of knowledge sources and the protection of data privacy, encompassing human-centered evaluation and oversight, such as bias identification, safety monitoring, and the assurance of cross-linguistic fairness. Recognizing these issues is crucial for promoting the responsible application of RAG in clinical practice and advancing its role in improving the quality of global health care services.[40]

## Results

We retrieved a total of 3,980 study records from PubMed, Embase, Web of Science, and Scopus. After deduplication and screening of titles, abstracts, and full texts, 248 studies met the inclusion criteria. Additionally, 3 relevant studies were manually identified, resulting in a total of 251 studies being analyzed. The PRISMA flow diagram is shown in Figure 2.

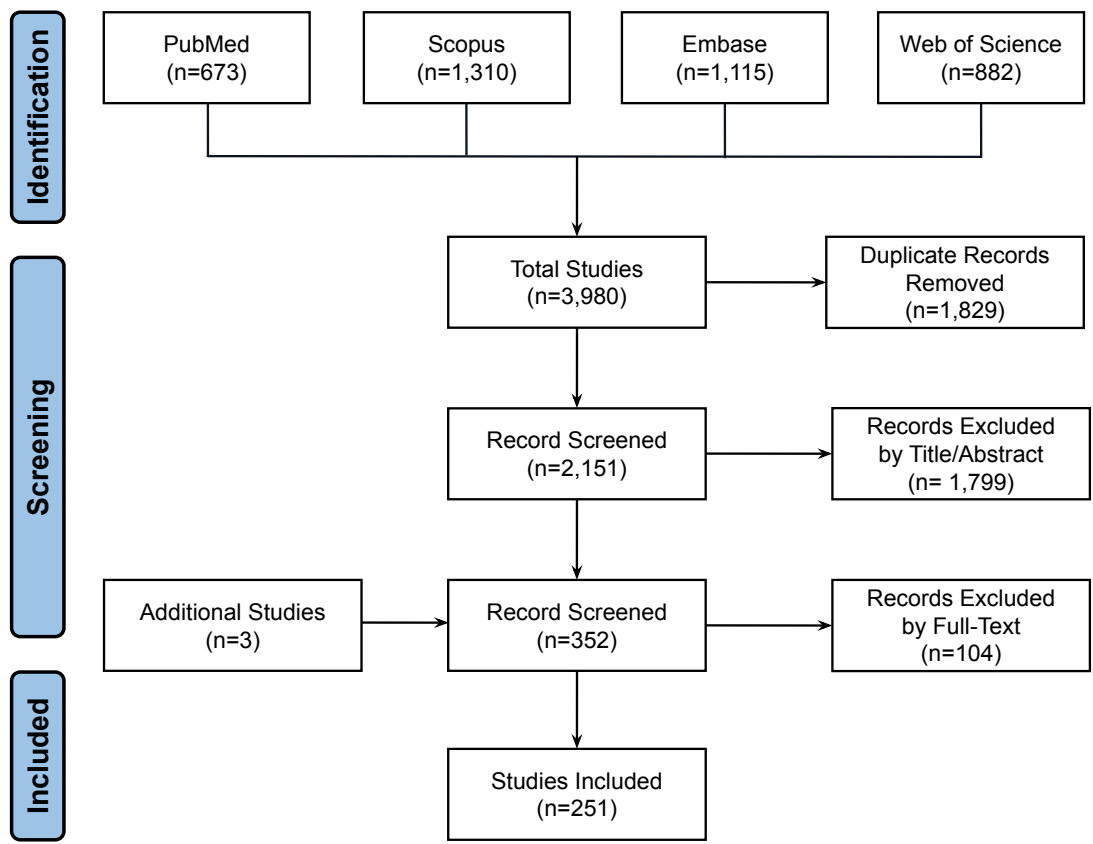

**Figure 2. PRISMA flow diagram for identifying related studies.** Our search retrieved 3,980 study records (n=673, 16.91% from PubMed; n=1,310, 32.91% from Scopus; n=1,115, 28.02% from Embase; n=882, 22.16% from Web of Science); of these, 2,151 (54.05%) were retained after deduplication. Following title, abstract, and full-text screening, 248 studies were included, along with 3 additional studies manually identified as relevant, resulting in a total of 251 included studies.

**Components of RAG Framework**

*External Retrieval Data*

In terms of data source (Figure 3 - Data Source), most studies (80.35%, 184/229) relied on publicly available data, while only 36 studies (15.72%) used private data, and 9 studies (3.93%) used both public and private data. This distribution pattern indicates that RAG research in medicine predominantly depends on open resources, with limited utilization of proprietary clinical data.

Meanwhile, RAG research primarily relies on diverse biomedical and clinical resources. As shown in Figure 3 (Data Type), biomedical scientific corpora constitute the main data source, used in 90 studies, with PubMed being a typical example. Clinical guidelines follow closely, adopted in 68 studies and underscoring the key role of evidence-based medicine in RAG applications. Online information (39 studies) and electronic health records (34 studies) are also common; the former reflects the demand for continuously updated medical knowledge, while the latter represents data support closely tied to real clinical practice. In addition, medical textbooks are incorporated in 31 studies as foundational learning and reference materials; knowledge graphs (30 studies), as a structured form of knowledge representation, offer different possibilities for medical knowledge retrieval; and custom-built datasets (30 studies) reflect researchers' exploration of constructing tailored resources for specific contexts. Overall, the data types involved in medical RAG encompass both structured and unstructured information, forming a diversified support framework.

*Retrieval Method*

As shown in Figure 3 (Retrieval Method), dense retrieval dominated among the included studies, with 189 studies (84.38%) adopting this approach. In contrast, sparse retrieval was used in only 11 studies (4.91%), while hybrid approaches were applied in 24 studies (10.71%). Dense retrieval methods were primarily implemented in two ways: one relied on general embedding models, such as OpenAI's text-embedding series;[47] the other employed medical-specific embedding models including BioBERT[48] and MedCPT,[49] which are trained on biomedical corpora to better capture the semantics of the field. However, it is noteworthy that most dense retrieval embedding models are still primarily trained on English corpora, posing significant limitations in non-English medical scenarios.[50] This is particularly evident when handling multilingual medical data, localized clinical guidelines, and non-English patient records, where semantic misinterpretation and reduced retrieval accuracy may occur.[40] Sparse retrieval methods mainly relied on the traditional BM25 algorithm,[51] but their usage frequency within the RAG framework is noticeably low. Hybrid retrieval approaches, which combine sparse and dense methods to balance lexical matching and semantic representation, exhibited certain advantages in some studies.[52]

*Generative LLM*

As shown in Figure 3 (LLM Type), proprietary LLMs were the most widely used, appearing in 103 studies (42.39%). Open-weight LLMs were utilized in 85 studies (34.98%), and a combination of both in 55 studies (22.63%). Proprietary LLMs were primarily from the OpenAI's GPT series,[10,11,53] while open-weight LLMs mainly include the DeepSeek, Gemma, LLaMA, and Qwen series.[14–25] Notably, medical-specific LLMs were rarely applied in existing RAG research. One possible reason is that proprietary medical LLMs (e.g., Med-Gemini[29,30]) have not released APIs to the public, limiting their accessibility for research and practice. Additionally, open-weight medical LLMs have developed more slowly than general LLMs, lagging in both scale and performance.[54]

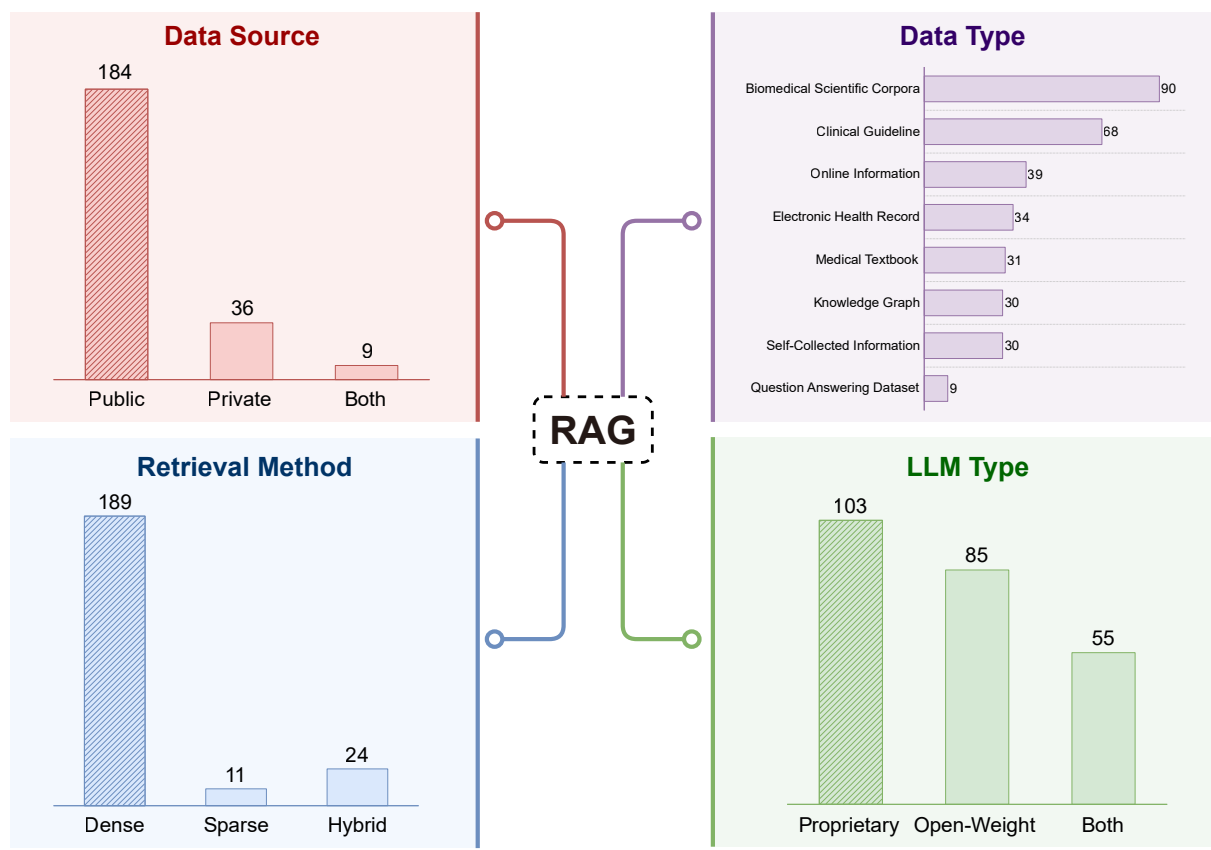

**Figure 3: Distribution of external retrieval data, retrieval methods, and generative LLMs among the 251 studies.** In terms of data source, 184 studies (80.35%) used public data, 36 studies (15.72%) used private data, and 9 studies (3.93%) used both. In terms of data type, the studies mainly employed biomedical scientific corpora (90 studies), clinical guideline (68 studies), online information (39 studies), electronic health record (34 studies), medical textbook (31 studies), knowledge graph (30 studies), self-collected information (30 studies), and question

answering dataset (9 studies). As for retrieval method, 189 studies (84.38%) applied dense retrieval, 11 studies (4.91%) applied sparse retrieval, and 24 studies (10.71%) applied hybrid retrieval. Regarding LLM type, 103 studies (42.39%) applied proprietary LLMs, 85 studies (34.98%) applied open-weight LLMs, and 55 studies (22.63%) applied both. It should be noted that only information explicitly reported in the studies is included here, while some studies did not provide detailed descriptions.

**Medical Specialty and Application Scenario**

*Medical Specialty Distribution*

The distribution of RAG across medical specialties exhibits considerable variability. Internal Medicine occupies the most prominent position, with 48 studies, reflecting its broad demand for synthesizing extensive clinical knowledge. Beyond this, Psychiatry and Neurology (24 studies), along with Radiology (22 studies), have shown a certain level of research activity. Additional specialties, including Preventive Medicine, Emergency Medicine, Orthopaedic Surgery, Medical Genetics and Genomics, and Obstetrics and Gynecology, have been explored to a lesser extent, while engagement in remaining specialties is limited. In general, the adoption of RAG across medical specialties remains uneven.

*Clinical Application Scenario*

Meanwhile, RAG has been explored in a variety of medical scenarios. Among them, medical question answering is the most widely used task, which supports clinicians with evidence retrieval, diagnostic reasoning, and clinical decision-making, while also assisting patients in obtaining understandable answers to medical inquiries.[55] Report generation is another application, where structured or semi-structured clinical data are automatically transformed into complete clinical reports (e.g., radiology or pathology reports) to reduce documentation burden.[55] Additionally, text summarization and information extraction have gained attention: the former condenses lengthy narratives to enhance information accessibility, while the latter converts unstructured text into structured data to support downstream analysis.[55] Other tasks, such as text simplification, have also been explored. Collectively, these tasks demonstrate the multifaceted potential of RAG in supporting clinical workflows.

**Evaluation and Ethical Considerations**

*Automatic and Human Evaluation*

In medical applications of RAG, evaluation methods demonstrate a relatively balanced distribution between automated and human evaluations, as shown in Figure 4 (Automatic and Human Evaluation). Automated evaluation was the most common setting, adopted in 112 studies (47.66%), while 43 studies (18.30%) only relied on human evaluation, and 80 studies (34.04%) combined both evaluations. Automated metrics generally fall into two categories: (i) text generation quality metrics (e.g., ROUGE,[56] BERTScore,[57] BLEU,[58] METEOR[59]), which assess linguistic quality and semantic similarity between generated content and reference texts; and (ii) task-specific performance metrics (e.g., accuracy, recall, F1 score, AUROC), typically applied in tasks such as multiple-choice question answering and clinical risk prediction. Human evaluation, by contrast, plays an indispensable role in dimensions not fully captured by automated metrics, with a primary focus on the factual accuracy and clinical utility of generated content, including completeness, relevance, fluency, as well as the detection of hallucination, bias, and safety concerns. Overall, current evaluation practices indicate the importance of balancing quantifiable performance metrics with subjective judgments of clinical acceptability.

*Ethical and Contextual Consideration*

As shown in Figure 4 (Ethical and Contextual Considerations), only 7 (2.79%) studies explicitly examined bias, 24 (9.56%) studies addressed safety, and 6 (2.39%) studies focused on low-resource settings. For bias assessment, most studies relied on small-scale expert validation. Some studies invited clinicians to rate bias in LLM outputs, while others directly examined differences across age, sex, race, and socioeconomic status. For safety evaluation, existing methods included detecting hallucinations and potential harms in generated responses, or applying rule-based safety filters. Although these explorations are valuable, they covered only a small fraction of the studies. In addition, research concerning low-resource settings involved countries such as India,[60,61] but overall remained very limited. These findings point to an imbalance in current evaluation practices, with insufficient attention given to ethical considerations.

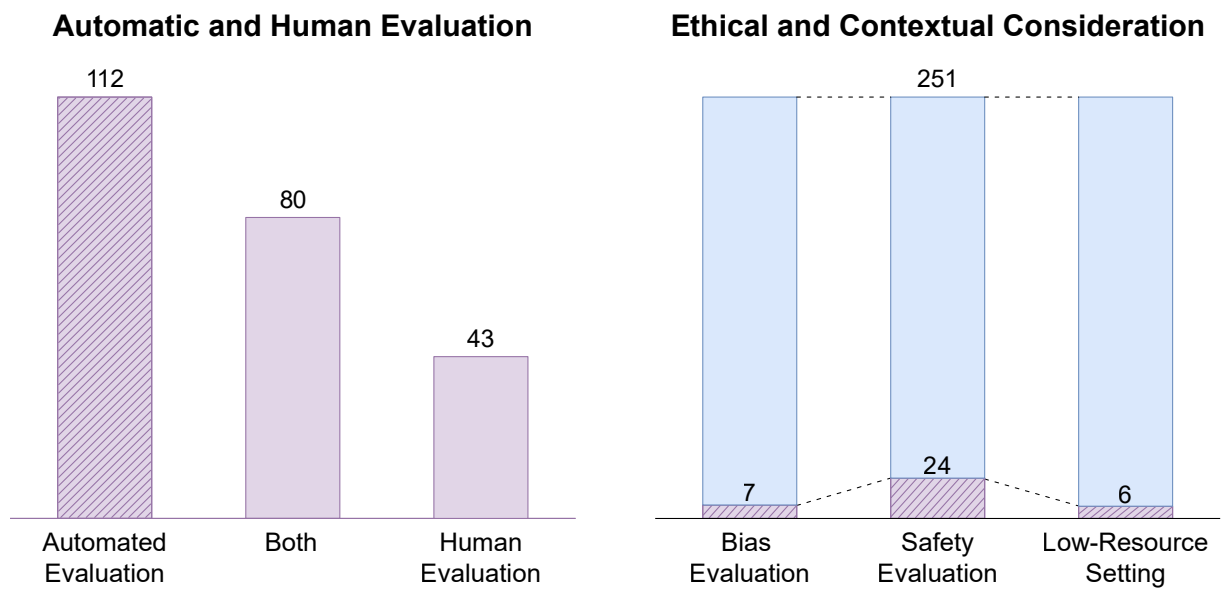

**Figure 4: Distribution of evaluation methods and ethical considerations among the 251 studies.** A total of 235 included studies reported evaluation methods, with automated evaluation (n = 112, 47.66%), human evaluation (n = 43, 18.30%), and a combination of both (n = 80, 34.04%). Among all 251 studies, only 7 studies explicitly examined bias, 24 studies addressed safety, and 6 studies were conducted in low-resource settings.

## Discussion

This scoping review systematically analyzed 251 RAG studies in medicine, revealing multifaceted characteristics and trends. Regarding data utilization, most studies relied on publicly available resources, with biomedical scientific corpora, clinical guidelines, and online information being the most common; real-world clinical data such as electronic health records were used only to a limited extent, while some studies also adopted medical textbooks and knowledge graphs. In terms of retrieval methods, dense retrieval emerged as the predominant approach, while sparse and hybrid strategies reported less frequently. For the generative model, proprietary LLMs dominated, with open-weight LLMs also used to some extent; however, LLMs specialized for medicine were rarely employed. Regarding specialty distribution, research was primarily concentrated in Internal Medicine, with other specialties also explored, though the overall distribution remained uneven. As for applications, medical question answering emerged as the main task. Other important applications include report generation, text summarization, and information extraction, all of which primarily focus on reducing the workload of clinicians. In terms of evaluation methods, automated evaluation was the main strategy, though many studies incorporated human evaluation to compensate for the inadequacies of automated metrics in clinical contexts. Notably, only a small number of studies considered bias, safety, or low-resource settings. Overall, while medical RAG research has established a certain foundation in methodological exploration, it remains in early stages regarding data availability, clinical validation, and responsible application.

The mode of data source selection reveals the constraints faced by the development of medical RAG. While the widespread use of open data has facilitated technical validation, it may also confine the application of RAG systems to the level of general medical knowledge, creating limitations in personalized health care scenarios where integration of real-world clinical data is essential. The limited integration of private data is primarily constrained by strict data privacy protection requirements, the lack of robust cross-institutional collaboration mechanisms, and the complexity of the implementation process. These challenges may hinder the further development of RAG technology in personalized medicine.

In terms of retrieval methods, the dominance of dense retrieval in medical RAG applications reflects its advantage in capturing medical semantic relations, particularly with the use of medical-specific embedding models such as MedCPT,[49] which shows the importance of domain adaptation for retrieval performance. However, the linguistic limitations faced by current dense retrieval methods are especially prominent. First, reliance on English-centric embedding models may hinder effective coverage of non-English medical data, thereby limiting the broader application of RAG technologies in global health care systems.[62,63] Second, such linguistic bias may exacerbate inequities in medical research and clinical practice, especially in low-resource languages and regions, where relevant medical knowledge is less likely to be adequately incorporated and utilized.[64] Although some studies have explored combining sparse or hybrid retrieval strategies to expand coverage, significant gaps remain in cross-lingual and fairness-related aspects of medical RAG, underscoring the need for further exploration in multilingual embeddings and localized adaptation.

A noteworthy phenomenon in the application of generative LLMs is the insufficient use of medical-specific LLMs. This may stem from the limited availability and relatively slower development progress of such models. While general LLMs perform well in text generation, they may face adaptability challenges in understanding complex medical terminology, clinical reasoning logic, and the integration of domain-specific knowledge. This gap between technical capability and application requirements warrants further attention.

The distribution of RAG applications in medicine reflects varying levels of acceptance and practical demands across task types. Applications related to question answering dominate; these applications essentially work by retrieving external knowledge to support medical knowledge access, diagnostic reasoning, and clinical decision-making. However, this category carries substantial risks as well: if retrieval sources are opaque, evidence is not verified, or generated content is not validated, clinicians may be misled into making incorrect judgments, and even pass harmful information on to patients—ultimately compromising diagnostic patient safety and patient trust. In contrast, tasks related to information processing (such as report generation, text summarization, and information extraction) carry comparatively lower clinical risks, making them an ideal entry point for RAG in clinical settings. Their primary value lies in reducing documentation burdens and improving information management efficiency, rather than intervening directly in clinical workflow.

The evaluation methods for current medical applications of RAG exhibit a diverse pattern, with the combined use of automated and human evaluation reflecting the complexity of medical AI evaluation. However, this diversity brings challenges in terms of result comparability and reproducibility. The use of text generation quality metrics and task-specific performance metrics illustrates the varying evaluation priorities across different medical tasks. Nevertheless, these metrics often fall short in capturing the full scope of clinical practicality and potential risks within medical contexts. At the same time, while human evaluation can compensate for some of these shortcomings, it is limited by subjectivity and cost, posing difficulties to scalability on a large-scale. More importantly, existing studies show limited attention to issues of safety, bias, and fairness, which may pose potential risks in clinical practice. The absence of robust mechanisms for detecting and mitigating these issues could undermine fairness in health care delivery and threaten patient safety. Moreover, insufficient research in low-resource settings highlights the broader challenges that medical RAG faces in advancing global health equity.

To advance RAG toward real clinical implementation, breakthroughs are needed across multiple critical dimensions, as shown in Figure 5. First, beyond continuous improvements in technical reliability, there is an urgent need for rigorous clinical validation to ensure that generated content is not only factually accurate but also clinically actionable. Second, it is equally critical to establish traceability and transparency mechanisms that enable clinicians to examine the retrieval sources and reasoning processes behind the output. Meanwhile, corresponding regulatory frameworks and ethical guidelines must be developed in parallel to ensure patient safety and enhance trust between health care providers and patients. Finally, it is important to recognize that the global deployment of medical RAG will encounter multidimensional challenges arising from linguistic, cultural, resource, and institutional disparities. Therefore, substantial progress in cross-linguistic and cross-cultural adaptation, as well as the assurance of fairness in low-resource settings, will be required before international implementation. Only by addressing these challenges can RAG achieve safe, trustworthy, and responsible clinical use on a global scale.

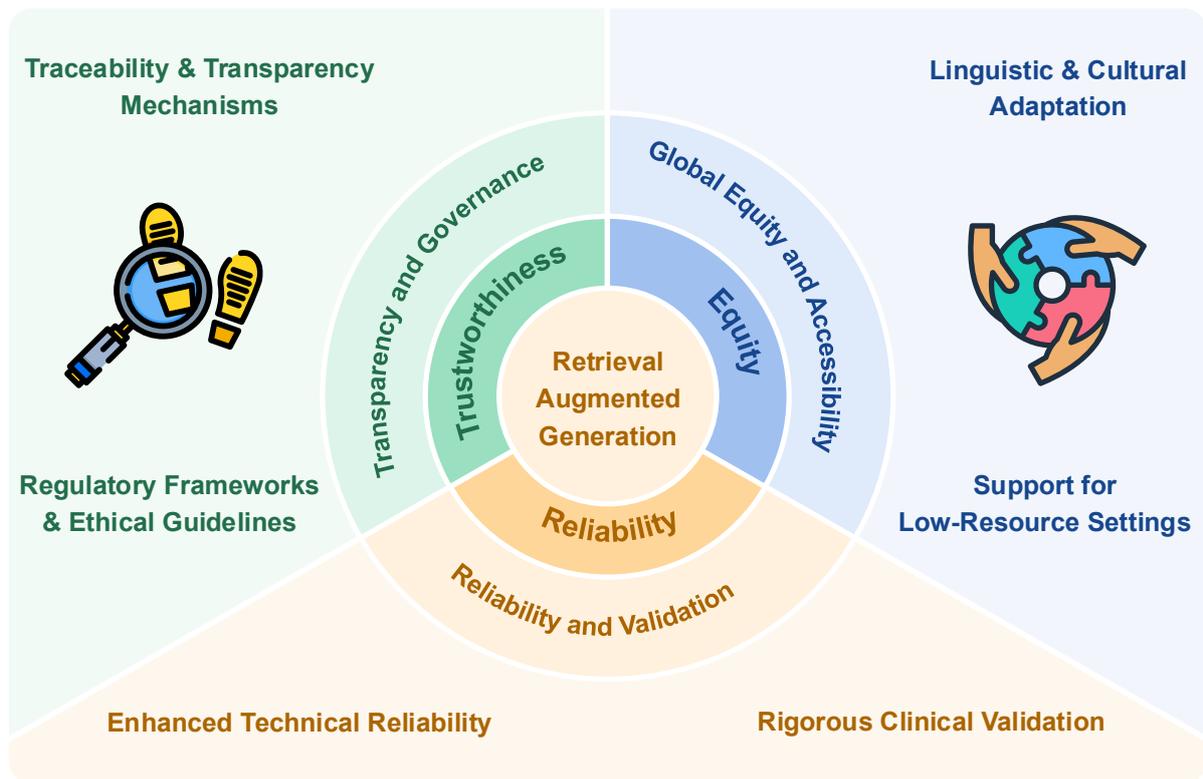

**Figure 5. Important directions for advancing medical RAG toward real clinical implementation.** This figure outlines three strategic directions required for responsible deployment of RAG in medicine: (1) Reliability—ensuring technical robustness and rigorous clinical validation to guarantee system-level stability and clinical actionability; (2) Trustworthiness—establishing traceability and transparency mechanisms alongside regulatory frameworks and ethical guidelines; (3) Equity—promoting cross-linguistic and cross-cultural adaptation while supporting low-resource settings to reduce global health disparities.

**Limitations**

This study has several limitations. First, we excluded non-English publications, which may introduce bias into the overview of RAG research in medicine. The results might differ if other languages and country-specific databases had been included. However, it is practically impossible to comprehensively cover all languages and national databases; therefore, we followed common practice in the domain. Second, artificial intelligence (AI) research is typically published in peer-reviewed conference proceedings, which are not well covered by existing search databases. Aside from Google Scholar, no other search engine can comprehensively capture all relevant AI studies, and Google Scholar itself has issues of reproducibility and search reliability. Lastly, given the extremely rapid development of this area, some recent relevant studies were inevitably missed during our search.

## Acknowledgments

This work was supported by the Duke-NUS Signature Research Program funded by the Ministry of Health, Singapore. Any opinions, findings and conclusions or recommendations expressed in this material are those of the author(s) and do not reflect the views of the Ministry of Health.

## Author Contributions

R.Y. and N.L. contributed to the conceptualization and methodology design. R.Y., M.W., H.L., X.L., W.Z., J.L., K.Y., and J.C.K.L. contributed to data curation. R.Y., H.L., Y.C., and W.X. contributed to visualization. R.Y., M.W., H.L., and Y.K. drafted the manuscript, with further improvements by J.C.L.O., D.T., C.H., D.S.W.T., and N.L. N.L. supervised the study. All authors contributed to the revision of the manuscript and approved the final version.

## Declaration of Interests

The authors declare no competing interests.

## Main Figure Titles and Legends

**Figure 1. Naive RAG framework.** The framework consists of three stages: first, the retrieval module obtains information relevant to the query; next, the query and the retrieved information are provided to the LLM; finally, the LLM generates the answer.

**Figure 2. PRISMA flow diagram for identifying related studies.** Our search retrieved 3,980 study records (n=673, 16.91% from PubMed; n=1,310, 32.91% from Scopus; n=1,115, 28.02% from Embase; n=882, 22.16% from Web of Science); of these, 2,151 (54.05%) were retained after deduplication. Following title, abstract, and full-text screening, 248 studies were included, along with 3 additional studies manually identified as relevant, resulting in a total of 251 included studies.

**Figure 3: Distribution of external retrieval data, retrieval methods, and generative LLMs among the 251 studies.** In terms of data source, 184 studies (80.35%) used public data, 36 studies (15.72%) used private data, and 9 studies (3.93%) used both. In terms of data type, the studies mainly employed biomedical scientific corpora (90 studies), clinical guideline (68 studies), online information (39 studies), electronic health record (34 studies), medical textbook (31 studies), knowledge graph (30 studies), self-collected information (30 studies), and question answering dataset (9 studies). As for retrieval method, 189 studies (84.38%) applied dense retrieval, 11 studies (4.91%) applied sparse retrieval, and 24 studies (10.71%) applied hybrid retrieval. Regarding LLM type, 103 studies (42.39%) applied proprietary LLMs, 85 studies (34.98%) applied open-weight LLMs, and 55 studies (22.63%) applied both. It should be noted that only information explicitly reported in the studies is included here, while some studies did not provide detailed descriptions.

**Figure 4: Distribution of evaluation methods and ethical considerations among the 251 studies.** A total of 235 included studies reported evaluation methods, with automated evaluation (n = 112, 47.66%), human evaluation (n = 43, 18.30%), and a combination of both (n = 80, 34.04%). Among all 251 studies, only 7 studies explicitly examined bias, 24 studies addressed safety, and 6 studies were conducted in low-resource settings.

**Figure 5. Important directions for advancing medical RAG toward real clinical implementation.** This figure outlines three strategic directions required for responsible deployment of RAG in medicine: (1) Reliability—ensuring technical robustness and rigorous clinical validation to guarantee system-level stability and clinical actionability; (2) Trustworthiness—establishing traceability and transparency mechanisms alongside regulatory frameworks and ethical guidelines; (3) Equity—promoting cross-linguistic and cross-cultural adaptation while supporting low-resource settings to reduce global health disparities.

## STAR★Methods

**Resource Availability**

*Lead Contact*

Requests for resources, data, and materials should be directed to, and will be fulfilled by, the Lead Contact, Nan Liu (email: liu.nan@duke-nus.edu.sg).

*Materials Availability*

This study did not generate new unique materials or reagents.

*Data and Code Availability*

Any data and code are available from the Lead Contact upon reasonable request.

**Search Strategy**

We conducted this scoping review following the Preferred Reporting Items for Systematic Reviews and Meta-Analyses Extension for Scoping Reviews (PRISMA-ScR) guideline to ensure transparency and rigor throughout the process.[65] We performed a comprehensive search across PubMed, Embase, Web of Science, and Scopus. The search strategy involved using a combination of terms related to "RAG" and "Medicine" to ensure maximum coverage and relevance. Our search period spans from 2017 (when the "Transformer" architecture was introduced[66]) to July 1st, 2025, to ensure comprehensive coverage of the full development trajectory of RAG technologies in the context of LLMs. Additionally, some important studies that could not be retrieved but were considered highly relevant were manually included. For the detailed search strategy, please refer to the Supplementary Material.

**Inclusion Criteria and Exclusion Criteria**

Studies were included if they met the following criteria: (1) Implemented a RAG framework, defined as leveraging retrieved external knowledge to improve the generation of LLMs; and (2) Were applied in medical scenarios, with potential impact and contributions to health care services. Additionally, we excluded studies that lacked abstracts, were non-English, were not peer-reviewed, were non-research types (such as review or perspective), as well as studies without clear RAG implementation details.

In the title and abstract screening stage, each study was independently screened by at least two researchers (RY, XL, KY, HL, JL, WZ, JL and MW) to determine eligibility. During the full-text screening stage, at least two researchers again independently screened the studies and extracted data. Any disagreements arising during this process were resolved through consultation with domain experts (RY, WX).

**Data Extraction**

For each included study, we extracted information across the following 6 perspectives: (1) metadata; (2) external retrieval data: data source, data availability, data scale, retrieval method; (3) model: generative model, model accessibility, embedding model; (4) application: medical specialty, application scenario; (5) evaluation: evaluation metrics, inclusion of human evaluation; (6) bias, safety, and low-resource setting: inclusion of bias evaluation, inclusion of safety evaluation, and application in low-resource settings.

Requests for resources should be directed to and will be fulfilled by the lead contact, Nan Liu (email: liu.nan@duke-nus.edu.sg).

# References

1. Shortliffe, E.H., and Sepúlveda, M.J. (2018). Clinical Decision Support in the Era of Artificial Intelligence. JAMA *320*, 2199–2200.

2. Jameson, J.L., and Longo, D.L. (2015). Precision medicine--personalized, problematic, and promising. N Engl J Med *372*, 2229–2234.

3. Anthropic. Claude 2. Available at: https://www.anthropic.com/news/claude-2. Accessed 1 November 2025.

4. Anthropic. Claude 3.5 Sonnet. Available at: https://www.anthropic.com/news/claude-3-5-sonnet. Accessed 1 November 2025.

5. Anthropic. Introducing Claude 4. Available at: https://www.anthropic.com/news/claude-4. Accessed 1 November 2025.

6. Anthropic. Claude Opus 4.1. Available at: https://www.anthropic.com/news/claude-opus-4-1. Accessed 1 November 2025.

7. Gemini Team, Anil, R., Borgeaud, S., Alayrac, J.-B., Yu, J., Soricut, R., Schalkwyk, J., Dai, A.M., Hauth, A., Millican, K., et al. (2023). Gemini: A family of highly capable multimodal models. arXiv [cs.CL]. https://doi.org/10.48550/ARXIV.2312.11805.

8. Gemini Team, Georgiev, P., Lei, V.I., Burnell, R., Bai, L., Gulati, A., Tanzer, G., Vincent, D., Pan, Z., Wang, S., et al. (2024). Gemini 1.5: Unlocking multimodal understanding across millions of tokens of context. arXiv [cs.CL]. https://doi.org/10.48550/ARXIV.2403.05530.

9. Comanici, G., Bieber, E., Schaekermann, M., Pasupat, I., Sachdeva, N., Dhillon, I., Blistein, M., Ram, O., Zhang, D., Rosen, E., et al. (2025). Gemini 2.5: Pushing the frontier with advanced reasoning, multimodality, long context, and next generation agentic capabilities. arXiv [cs.CL]. https://doi.org/10.48550/ARXIV.2507.06261.

10. Brown, T.B., Mann, B., Ryder, N., Subbiah, M., Kaplan, J., Dhariwal, P., Neelakantan, A., Shyam, P., Sastry, G., Askell, A., et al. (2020). Language Models are Few-Shot Learners. arXiv [cs.CL]. https://doi.org/10.48550/ARXIV.2005.14165.

11. OpenAI, Achiam, J., Adler, S., Agarwal, S., Ahmad, L., Akkaya, I., Aleman, F.L., Almeida, D., Altenschmidt, J., Altman, S., et al. (2023). GPT-4 Technical Report. arXiv [cs.CL]. https://doi.org/10.48550/ARXIV.2303.08774.

12. OpenAI, :, Hurst, A., Lerer, A., Goucher, A.P., Perelman, A., Ramesh, A., Clark, A., Ostrow, A.J., Welihinda, A., et al. (2024). GPT-4o System Card. arXiv [cs.CL]. https://doi.org/10.48550/ARXIV.2410.21276.

13. OpenAI. GPT-5 System Card. Available at: https://openai.com/index/gpt-5-system-card/. Accessed 1 November 2025.

14. DeepSeek-AI, Liu, A., Feng, B., Xue, B., Wang, B., Wu, B., Lu, C., Zhao, C., Deng, C., Zhang, C., et al. (2024). DeepSeek-V3 Technical Report. arXiv [cs.CL]. https://doi.org/10.48550/ARXIV.2412.19437.

15. Gemma Team, Mesnard, T., Hardin, C., Dadashi, R., Bhupatiraju, S., Pathak, S., Sifre, L., Rivière, M., Kale, M.S., Love, J., et al. (2024). Gemma: Open models based on Gemini research and technology. arXiv [cs.CL]. https://doi.org/10.48550/ARXIV.2403.08295.

16. Gemma Team, Riviere, M., Pathak, S., Sessa, P.G., Hardin, C., Bhupatiraju, S., Hussenot, L., Mesnard, T., Shahriari, B., Ramé, A., et al. (2024). Gemma 2: Improving open language models at a practical size. arXiv [cs.CL]. https://doi.org/10.48550/ARXIV.2408.00118.

17. Gemma Team, Kamath, A., Ferret, J., Pathak, S., Vieillard, N., Merhej, R., Perrin, S., Matejovicova, T., Ramé, A., Rivière, M., et al. (2025). Gemma 3 Technical Report. arXiv [cs.CL]. https://doi.org/10.48550/ARXIV.2503.19786.

18. Touvron, H., Lavril, T., Izacard, G., Martinet, X., Lachaux, M.-A., Lacroix, T., Rozière, B., Goyal, N., Hambro, E., Azhar, F., et al. (2023). LLaMA: Open and efficient foundation language models. arXiv [cs.CL]. https://doi.org/10.48550/ARXIV.2302.13971.

19. Grattafiori, A., Dubey, A., Jauhri, A., Pandey, A., Kadian, A., Al-Dahle, A., Letman, A., Mathur, A., Schelten, A., Vaughan, A., et al. (2024). The Llama 3 herd of models. arXiv [cs.AI]. https://doi.org/10.48550/ARXIV.2407.21783.

20. Meta AI. The Llama 4 herd: The beginning of a new era of natively multimodal AI innovation. Available at: https://ai.meta.com/blog/llama-4-multimodal-intelligence/. Accessed 1 November 2025.

21. Bai, J., Bai, S., Chu, Y., Cui, Z., Dang, K., Deng, X., Fan, Y., Ge, W., Han, Y., Huang, F., et al. (2023). Qwen Technical Report. arXiv [cs.CL]. https://doi.org/10.48550/ARXIV.2309.16609.

22. Yang, A., Yang, B., Hui, B., Zheng, B., Yu, B., Zhou, C., Li, C., Li, C., Liu, D., Huang, F., et al. (2024). Qwen2 Technical Report. arXiv [cs.CL]. https://doi.org/10.48550/ARXIV.2407.10671.

23. Qwen, :, Yang, A., Yang, B., Zhang, B., Hui, B., Zheng, B., Yu, B., Li, C., Liu, D., et al. (2024). Qwen2.5 Technical Report. arXiv [cs.CL]. https://doi.org/10.48550/ARXIV.2412.15115.

24. Yang, A., Li, A., Yang, B., Zhang, B., Hui, B., Zheng, B., Yu, B., Gao, C., Huang, C., Lv, C., et al. (2025). Qwen3 Technical Report. arXiv [cs.CL]. https://doi.org/10.48550/ARXIV.2505.09388.

25. DeepSeek-AI, Guo, D., Yang, D., Zhang, H., Song, J., Zhang, R., Xu, R., Zhu, Q., Ma, S., Wang, P., et al. (2025). DeepSeek-R1: Incentivizing Reasoning Capability in LLMs via Reinforcement Learning. arXiv [cs.CL]. https://doi.org/10.48550/ARXIV.2501.12948.

26. OpenAI, :, Jaech, A., Kalai, A., Lerer, A., Richardson, A., El-Kishky, A., Low, A., Helyar, A., Madry, A., et al. (2024). OpenAI o1 System Card. arXiv [cs.AI]. https://doi.org/10.48550/ARXIV.2412.16720.

27. OpenAI. OpenAI o3-mini System Card. Available at: https://openai.com/index/o3-mini-system-card/. Accessed 1 November 2025.

28. OpenAI. OpenAI o3 and o4-mini System Card. Available at: https://openai.com/index/o3-o4-mini-system-card/. Accessed 1 November 2025.

29. Saab, K., Tu, T., Weng, W.-H., Tanno, R., Stutz, D., Wulczyn, E., Zhang, F., Strother, T., Park, C., Vedadi, E., et al. (2024). Capabilities of Gemini models in medicine. arXiv [cs.AI]. https://doi.org/10.48550/ARXIV.2404.18416.

30. Yang, L., Xu, S., Sellergren, A., Kohlberger, T., Zhou, Y., Ktena, I., Kiraly, A., Ahmed, F., Hormozdiari, F., Jaroensri, T., et al. (2024). Advancing multimodal medical capabilities of Gemini. arXiv [cs.CV]. https://doi.org/10.48550/ARXIV.2405.03162.

31. Sellergren, A., Kazemzadeh, S., Jaroensri, T., Kiraly, A., Traverse, M., Kohlberger, T., Xu, S., Jamil, F., Hughes, C., Lau, C., et al. (2025). MedGemma Technical Report. arXiv [cs.AI]. https://doi.org/10.48550/ARXIV.2507.05201.

32. Thirunavukarasu, A.J., Ting, D.S.J., Elangovan, K., Gutierrez, L., Tan, T.F., and Ting, D.S.W. (2023). Large language models in medicine. Nat Med *29*, 1930–1940.

33. Yang, R., Tan, T.F., Lu, W., Thirunavukarasu, A.J., Ting, D.S.W., and Liu, N. (2023). Large language models in health care: Development, applications, and challenges. Health Care Sci *2*, 255–263.

34. McDuff, D., Schaekermann, M., Tu, T., Palepu, A., Wang, A., Garrison, J., Singhal, K., Sharma, Y., Azizi, S., Kulkarni, K., et al. (2025). Towards accurate differential diagnosis with large language models. Nature *642*, 451–457.

35. Vedadi, E., Barrett, D., Harris, N., Wulczyn, E., Reddy, S., Ruparel, R., Schaekermann, M., Strother, T., Tanno, R., Sharma, Y., et al. (2025). Towards physician-centered oversight of conversational diagnostic AI. arXiv [cs.AI]. https://doi.org/10.48550/ARXIV.2507.15743.

36. Yang, R., Liu, H., Marrese-Taylor, E., Zeng, Q., Ke, Y., Li, W., Cheng, L., Chen, Q., Caverlee, J., Matsuo, Y., et al. (2024). KG-rank: Enhancing large language models for medical QA with knowledge graphs and ranking techniques. In Proceedings of the 23rd Workshop on Biomedical Natural Language Processing (Association for Computational Linguistics), pp. 155–166.

37. Ke, Y., Yang, R., Lie, S.A., Lim, T.X.Y., Ning, Y., Li, I., Abdullah, H.R., Ting, D.S.W., and Liu, N. (2024). Mitigating Cognitive Biases in Clinical Decision-Making Through Multi-Agent Conversations Using Large Language Models: Simulation Study. J Med Internet Res *26*, e59439.

38. Abd-Alrazaq, A., AlSaad, R., Alhuwail, D., Ahmed, A., Healy, P.M., Latifi, S., Aziz, S., Damseh, R., Alabed Alrazak, S., and Sheikh, J. (2023). Large Language Models in Medical Education: Opportunities, Challenges, and Future Directions. JMIR Med Educ *9*, e48291.

39. Yang, R., Tong, J., Wang, H., Huang, H., Hu, Z., Li, P., Liu, N., Lindsell, C.J., Pencina, M.J., Chen, Y., et al. (2025). Enabling inclusive systematic reviews: incorporating preprint articles with large language model-driven evaluations. J Am Med Inform Assoc.

40. Yang, R., Ning, Y., Keppo, E., Liu, M., Hong, C., Bitterman, D.S., Ong, J.C.L., Ting, D.S.W., and Liu, N. (2025). Retrieval-augmented generation for generative artificial intelligence in health care. Npj Health Syst. *2*.

41. Omiye, J.A., Lester, J.C., Spichak, S., Rotemberg, V., and Daneshjou, R. (2023). Large language models propagate race-based medicine. NPJ Digit Med *6*, 195.

42. Ong, J.C.L., Chang, S.Y.-H., William, W., Butte, A.J., Shah, N.H., Chew, L.S.T., Liu, N., Doshi-Velez, F., Lu, W., Savulescu, J., et al. (2024). Ethical and regulatory challenges of large language models in medicine. Lancet Digit Health *6*, e428–e432.

43. He, J., Zhang, B., Rouhizadeh, H., Chen, Y., Yang, R., Lu, J., Chen, X., Liu, N., Li, I., and Teodoro, D. (2025). Retrieval-Augmented Generation in biomedicine: A survey of technologies, datasets, and clinical applications. arXiv [q-bio.OT]. https://doi.org/10.48550/ARXIV.2505.01146.

44. Gao, Y., Xiong, Y., Gao, X., Jia, K., Pan, J., Bi, Y., Dai, Y., Sun, J., Wang, M., and Wang, H. (2023). Retrieval-Augmented Generation for Large Language Models: A Survey.

45. Liu, S., McCoy, A.B., and Wright, A. (2025). Improving large language model applications in biomedicine with retrieval-augmented generation: a systematic review, meta-analysis, and clinical development guidelines. J Am Med Inform Assoc *32*, 605–615.

46. Amugongo, L.M., Mascheroni, P., Brooks, S., Doering, S., and Seidel, J. (2025).

Retrieval augmented generation for large language models in healthcare: A systematic review. PLOS Digital Health *4*, e0000877.

47. OpenAI. Vector embeddings. Available at: https://platform.openai.com/docs/guides/embeddings. Accessed 1 November 2025.

48. Lee, J., Yoon, W., Kim, S., Kim, D., Kim, S., So, C.H., and Kang, J. (2020). BioBERT: a pre-trained biomedical language representation model for biomedical text mining. Bioinformatics *36*, 1234–1240.

49. Jin, Q., Kim, W., Chen, Q., Comeau, D.C., Yeganova, L., Wilbur, W.J., and Lu, Z. (2023). MedCPT: Contrastive Pre-trained Transformers with large-scale PubMed search logs for zero-shot biomedical information retrieval. Bioinformatics *39*.

50. Xuan, W., Yang, R., Qi, H., Zeng, Q., Xiao, Y., Feng, A., Liu, D., Xing, Y., Wang, J., Gao, F., et al. (2025). MMLU-ProX: A multilingual benchmark for advanced large language model evaluation. arXiv [cs.CL]. https://doi.org/10.48550/ARXIV.2503.10497.

51. Robertson, S., and Zaragoza, H. (2009). The probabilistic relevance framework: BM25 and beyond. Found. Trends® Inf. Retr. *3*, 333–389.

52. Liu, H., Soroush, A., Nestor, J.G., Park, E., Idnay, B., Fang, Y., Pan, J., Liao, S., Bernard, M., Peng, Y., et al. (2024). Retrieval augmented scientific claim verification. Jamia Open *7*, ooae021.

53. OpenAI. Introducing ChatGPT. Available at: https://openai.com/index/chatgpt/. Accessed 1 November 2025.

54. Wu, J., Gu, B., Zhou, R., Xie, K., Snyder, D., Jiang, Y., Carducci, V., Wyss, R., Desai, R.J., Alsentzer, E., et al. (2025). BRIDGE: Benchmarking large language models for understanding real-world clinical practice text. arXiv [cs.CL]. https://doi.org/10.48550/ARXIV.2504.19467.

55. Yang, R., Zeng, Q., You, K., Qiao, Y., Huang, L., Hsieh, C.-C., Rosand, B., Goldwasser, J., Dave, A., Keenan, T., et al. (2024). Ascle-A Python Natural Language Processing Toolkit for Medical Text Generation: Development and Evaluation Study. J Med Internet Res *26*, e60601.

56. Lin, C.-Y. (2004). Rouge: A package for automatic evaluation of summaries.

57. Zhang, T., Kishore, V., Wu, F., Weinberger, K.Q., and Artzi, Y. (2019). BERTScore: Evaluating Text Generation with BERT. arXiv [cs.CL]. https://doi.org/10.48550/ARXIV.1904.09675.

58. Papineni, K., Roukos, S., Ward, T., and Zhu, W.-J. (2001). BLEU. In Proceedings of the

40th Annual Meeting on Association for Computational Linguistics - ACL '02 (Association for Computational Linguistics).

59. Banerjee, S., and Lavie, A. (2005). METEOR: An Automatic Metric for MT Evaluation with Improved Correlation with Human Judgments. In Proceedings of the ACL Workshop on Intrinsic and Extrinsic Evaluation Measures for Machine Translation and/or Summarization, pp. 65–72.

60. Al Ghadban, Y., Lu, H. (yvonne), Adavi, U., Sharma, A., Gara, S., Das, N., Kumar, B., John, R., Devarsetty, P., and Hirst, J.E. (2023). Transforming Healthcare Education: Harnessing Large Language Models for Frontline Health Worker Capacity Building using Retrieval-Augmented Generation. medRxiv, 2023.12.15.23300009. https://doi.org/10.1101/2023.12.15.23300009.

61. Sun H, Li Q, Wang J, et al. (2024). A RAG-based Medical Assistant Especially for Infectious Diseases. IEEE Explore.

62. Yang, R., Nair, S.V., Ke, Y., D'Agostino, D., Liu, M., Ning, Y., and Liu, N. (2024). Disparities in clinical studies of AI enabled applications from a global perspective. NPJ Digit Med *7*, 209.

63. Akbarialiabad, H., Sadeghian, N., Haghighat, S., Grada, A., Paydar, S., Haghighi, A., Kvedar, J.C., and Sewankambo, N.K. (2025). The utility of generative AI in advancing global health. NEJM AI *2*.

64. Localizing AI in the global south (2025). Nat. Mach. Intell.

65. Tricco, A.C., Lillie, E., Zarin, W., O'Brien, K.K., Colquhoun, H., Levac, D., Moher, D., Peters, M.D.J., Horsley, T., Weeks, L., et al. (2018). PRISMA Extension for Scoping Reviews (PRISMA-ScR): Checklist and Explanation. Ann Intern Med *169*, 467–473.

66. Vaswani, A., Shazeer, N., Parmar, N., Uszkoreit, J., Jones, L., Gomez, A.N., Kaiser, L., and Polosukhin, I. (2017). Attention is all you need. arXiv [cs.CL]. https://doi.org/10.48550/ARXIV.1706.03762.